\documentclass{llncs}

\usepackage{amsmath}
\usepackage[ruled,vlined]{algorithm2e}
\usepackage{subfigure}
\usepackage{tikz}
\usetikzlibrary{arrows, positioning}

\newcommand\shrink[1]{}
\newcommand\spara[1]{\vspace{2mm}\noindent {\bf #1}}

\def\xor{\oplus}
\def\iff{\Leftrightarrow}
\def\lar{\ensuremath{\leftarrow}}

\def\X{{\bf X}}
\def\x{{\bf x}}
\def\Y{{\bf Y}}
\def\y{{\bf y}}
\def\Z{{\bf Z}}
\def\z{{\bf z}}

\def\M{{\bf M}}
\def\R{{\bf R}}
\def\C{{\bf C}}

\def\true{{\sf true}}
\def\false{{\sf false}}

\def\NP{\mathsf{NP}}
\def\P{\mathsf{P}}

\def\cd{{\tt c2d}\xspace}
\def\cdf{{\tt c2d\_forget}\xspace}
\def\coproc{{\tt Coprocessor}\xspace}

\begin{document}

\title{On Compiling DNNFs without Determinism}

\author{Umut Oztok \and Adnan Darwiche}

\institute{Computer Science Department\\ University of California, Los Angeles\\ 
           Los Angeles, CA 90095, USA \\
\email{\{umut,darwiche\}@cs.ucla.edu}}

\maketitle
\bibliographystyle{splncs03}

\begin{abstract} 

  State-of-the-art knowledge compilers generate {\em deterministic} subsets of
  DNNF, which have been recently shown to be exponentially less succinct than
  DNNF. In this paper, we propose a new method to compile DNNFs without
  enforcing determinism necessarily. Our approach is based on compiling
  deterministic DNNFs with the addition of auxiliary variables to the input
  formula. These variables are then existentially quantified from the
  deterministic structure in linear time, which would lead to a DNNF that is
  equivalent to the input formula and not necessarily deterministic. On the
  theoretical side, we show that the new method could generate exponentially
  smaller DNNFs than deterministic ones, even by adding a single auxiliary
  variable. Further, we show that various existing techniques that introduce
  auxiliary variables to the input formulas can be employed in our framework. On
  the practical side, we empirically demonstrate that our new method can
  significantly advance DNNF compilation on certain benchmarks.
  

\end{abstract}

\section{Introduction}

Decomposability and determinism are two fundamental properties that underlie
many tractable representations in propositional logic. Decomposability is the
characteristic property of decomposable negation normal form
(DNNF)~\cite{Darwiche01}, and adding determinism to DNNF leads to deterministic
DNNF (d-DNNF)~\cite{Darwiche00b}, which includes many other representations,
such as sentential decision diagrams (SDDs)~\cite{Darwiche11} and ordered binary
decision diagrams (OBDDs)~\cite{Bryant86}. 

The key property of deterministic subsets of DNNF is their ability to render the
query of model counting tractable, which is key to probabilistic reasoning (see,
e.g., \cite{Roth96,Darwiche02b,ChaviraD08}). On the other hand, decomposability
without determinism is also sufficient to ensure the tractability of many
interesting queries, such as clausal entailment and cardinality minimization.
Indeed, these queries are enough for various applications, which do not require
efficient computation of model counting. For example, constructing DNNFs would
suffice to perform required reasoning tasks efficiently for model-based
diagnosis~(e.g., \cite{Darwiche00,Barrett05,HuangD05}) and testing (e.g.,
\cite{SchumannHS10,SchumannS10}).

However, state-of-the-art knowledge compilers all generate deterministic subsets
of DNNF (see, e.g., \cite{Darwiche04,MuiseMB12,OztokD15}). Yet, unsurprisingly,
the addition of determinism comes with a cost of generating less succinct
representations. In particular, as recently shown~\cite{BovaCMS16}, DNNF is
exponentially more succinct than its deterministic subsets. Therefore, for those
applications where only decomposability is sufficient, compiling a deterministic
subset of DNNF not only implies performing more work than necessary, but it
could also result in generating larger DNNFs which would make reasoning tasks
less efficient (if compilation is possible at all). Still, all existing
compilers that we know of to generate decomposability also ensure determinism.

In this paper, we focus on compiling DNNFs without enforcing determinism, and
make several contributions in that matter. Our main contribution is a new
methodology to compile DNNFs by leveraging existing knowledge compilers. The key
insight behind our approach is a new type of equivalence relation between two
Boolean functions: a Boolean function $f(\X)$ over variables $\X$ is {\em
equivalent modulo forgetting} to another Boolean function $g(\X,\Y)$ over
variables $\X$ and $\Y$ iff existentially quantifying (also known as,
forgetting) variables $\Y$ from $g$ results in a function equivalent to $f$. The
relevance of this notion to DNNF compilation is the well-known result that one
can forget arbitrarily many variables on a given DNNF in linear time in the DNNF
size, without losing the property of decomposability but not necessarily
determinism~\cite{Darwiche01}. Thus, instead of compiling function~$f$ directly,
one can compile function~$g$ into a deterministic DNNF using existing compilers,
on which forgetting variables $\Y$ would result in a DNNF that is not
necessarily deterministic and equivalent to~$f$.

The usefulness of our new approach depends on two important questions, which we
address in this paper both theoretically and empirically: (i) to what extend
forgetting variables could lead to more compact DNNFs without determinism than
deterministic DNNFs, and (ii) how can one identify functions that are equivalent
modulo forgetting. On the theoretical side, we present two main results. First,
we show that even forgetting a single auxiliary variable can lead to exponential
difference between sizes of DNNFs with and without determinism. Second, we study
various existing approaches, such as Tseitin transformation~\cite{Tseitin68},
extended resolution~\cite{Tseitin70}, and bounded variable addition
(BVA)~\cite{MantheyHB12}, where auxiliary variables are introduced to formulas,
mostly to obtain an equisatisfiable formula so that SAT task can be performed or
becomes easy.\footnote{Two formulas are equisatisfiable when the satisfiability
of one depends on the other.} We show that those existing techniques indeed
correspond to generating functions that are equivalent modulo forgetting, and
hence offering some practical ideas to apply to our approach. In particular, we
show that BVA would generate CNFs without increasing the treewidth of the input
CNF much in the worst case, and could potentially reduce it to a bounded value
from an unbounded value. Since CNF-to-DNNF compilation is tractable for bounded
treewidth~\cite{Darwiche01}, this result shows the potential of BVA on DNNF
compilation. On the practical side, we demonstrate that BVA, which turns out to
be useful for SAT solving, can significantly advance DNNF compilation.

This paper is structured as follows. We start with providing some technical
preliminaries in Section~\ref{sec:pre}. We then describe our new method in
detail in Section~\ref{sec:emf}. This is followed by showing that forgetting a
single auxiliary variable can lead to exponential separation between DNNFs with
and without determinism in Section~\ref{sec:forget}. We then make a treatment of
various existing approaches in the literature as equivalent modulo forgetting
transformations in Section~\ref{sec:emf_trans}. After providing an empirical
evaluation of our new approach in Section~\ref{sec:exp}, we continue with a
discussion on related work in Section~\ref{sec:related}. We conclude the paper
with a few remarks in Section~\ref{sec:conc}. 

\section{Technical Preliminaries}\label{sec:pre}

In this section, we will briefly introduce the concepts that will be used
throughout the paper. We will use upper-case letters (e.g., $X$) to denote
variables and lower-case letters (e.g., $x$) to denote their instantiations.
That is, $x$ is a {\em literal} denoting $X$ or $\neg X$. We will use bold
upper-case letters (e.g., $\X$) to denote sets of variables and bold lower-case
letters (e.g., $\x$) to denote their instantiations.

A {\em Boolean function} $f$ over variables $\Z$, denoted $f(\Z)$, is a function
that maps each instantiation $\z$ of variables $\Z$ to either 1/$\true$ or
0/$\false$.  A {\em trivial} Boolean function maps all its inputs to $\true$
(denoted~$\top$) or maps them all to $\false$ (denoted $\bot$).  An
instantiation $\z$ {\em satisfies} function $f$ iff $f$ maps $\z$ to $\true$.
In this case, $\z$ is said to be a {\em model} of function $f$. The {\em model
count} of function $f$ is the number of models of~$f$. Two functions $f$ and $g$
are logically equivalent, denoted $f \equiv g$, iff they have the same set of
models. The {\em conditioning} of function~$f$ on instantiation~$\x$, denoted
$f|\x$, is the sub-function obtained by setting variables~$\X$ to their values
in $\x$. The {\em existential quantification} of variable $X$ from function $f$,
denoted $\exists\,X.\, f$, is the function obtained by disjoining functions
$f|X$ and $f|\neg X$ (that is, $\exists\,X.\, f = f|X \vee f|\neg X$).
Existential quantification is also known as {\em forgetting}, and can also be
performed on a set of variables $\X$ by successively quantifying variables in
$\X$. We will combine Boolean functions using the traditional Boolean operators,
such as $\wedge$, $\vee$, $\xor$, and $\iff$.

\spara{CNF:} A {\em conjunctive normal form} (CNF) is a conjunction of clauses,
where each clause is a disjunction of literals. For instance, $(X \vee \neg Y)
\wedge (\neg X \vee Y \vee Z) \wedge \neg Z$ is a CNF with three clauses.
Conditioning CNF $\Delta$ on literal $\ell$ amounts to removing literal $\neg
\ell$ from all clauses and then dropping all clauses that contain literal
$\ell$. 

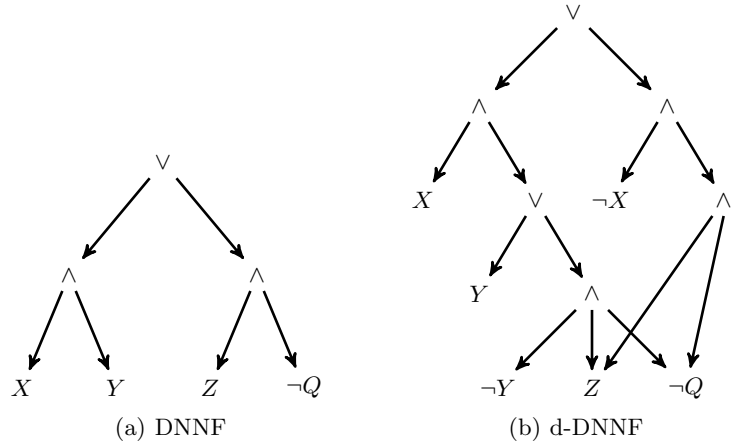
\begin{figure}[t]
  \centering
  \subfigure[DNNF]{
    \label{fig:dnnf_1}
    \begin{tikzpicture}[->, >=stealth', line width=1pt,
                   level 1/.style={sibling distance = 2.5cm,  level distance = 1.5cm},
                   level 2/.style={sibling distance = 1.25cm, level distance = 1.5cm}]

\tikzstyle{nnf} = [minimum size=.25cm]
\tikzset{every node/.style={nnf}}

\node {$\vee$}
  child{ 
    node {$\wedge$}
      child{node {$X$}}
      child{node {$Y$}}
  }
  child{ 
    node {$\wedge$}
      child{node {$Z$}}
      child{node {$\neg Q$}}
  };
\end{tikzpicture}
  }\qquad
  \subfigure[d-DNNF]{
    \label{fig:dnnf_2}
    \begin{tikzpicture}[->, >=stealth', line width=1pt,
                   level 1/.style={sibling distance = 2.5cm, level distance = 1.25cm},
                   level 2/.style={sibling distance = 1.5cm, level distance = 1.25cm},
                   level 3/.style={sibling distance = 1.5cm, level distance = 1.25cm},
                   level 4/.style={sibling distance = 1.25cm}]

\tikzstyle{nnf} = [minimum size=.25cm]
\tikzset{every node/.style={nnf}}

\node {$\vee$}
  child{ 
    node {$\wedge$}
      child{node {$X$}}
      child{
        node {$\vee$}
          child{
            node {$Y$}
          }
          child{
            node {$\wedge$}
              child{node {$\neg Y$}}
              child{node (z) {$Z$}}
              child{node (q) {$\neg Q$}}
          }
      }
  }
  child{
    node {$\wedge$}
      child{node {$\neg X$}}
      child{
        node (a) {$\wedge$} 
      }
  };

\draw [->] (a) -- (z);
\draw [->] (a.south) -- (q);
\end{tikzpicture}
  }
  \caption{A DNNF and a d-DNNF for the CNF $(X \vee \neg Q) \wedge (X \vee Z)
  \wedge (Y \vee \neg Q) \wedge (Y \vee Z)$.}
  \label{fig:dnnf}
\end{figure}

\spara{NNF:} A {\em negation normal form} (NNF) is a rooted, directed acyclic
graph whose internal nodes are labeled with either conjunctions (i.e., $\wedge$)
or disjunctions (i.e., $\vee$) and whose leaf nodes are labeled with either
literals or constants $\top$ and $\bot$~\cite{DarwicheM02}. A conjunction is
{\em decomposable} iff each pair of its conjuncts share no
variables~\cite{Darwiche01}. A disjunction is {\em deterministic} iff each pair
of its disjuncts are inconsistent with each other~\cite{Darwiche00b}. A {\em
decomposable negation normal form} (DNNF) is an NNF whose conjunctions are
decomposable~\cite{Darwiche01}. A {\em deterministic} DNNF (d-DNNF) is a DNNF
whose disjunctions are deterministic~\cite{Darwiche00b}. For instance,
Fig.~\ref{fig:dnnf} illustrates a DNNF and a d-DNNF that are both equivalent to
the CNF $(X \vee Z \wedge (X \vee \neg Q) \wedge (Y \vee Z) \wedge (Y \vee \neg
Q)$ (note that the former is not necessarily deterministic).

\section{Compiling DNNFs through Forgetting Variables}\label{sec:emf}

In this section, we will describe the proposed methodology, which is based on a
new type of equivalence relation between two functions. 

\begin{definition}
Let $f(\X)$ and $g(\X,\Y)$ be two Boolean functions, where variables $\X$ and
$\Y$ are disjoint. Then function $f$ is said to be \underline{equivalent modulo
forgetting} (emf) to function $g$ iff the following holds:
$$
f(\X) \equiv \exists\,\Y.\, g(\X,\Y).
$$
\end{definition}

\noindent Intuitively, the models of functions $f$ and $g$ match on their values
over variables $\X$. Specifically, for each model $\x$ of $f$, there must exist
an instantiation $\y$ such that $\x\y$ is a model of $g$. Similarly, for each
model $\x\y$ of $g$, $\x$ must be a model of $f$. In other words, function $f$
says everything function $g$ says on variables $\X$. Hence, variables $\Y$ only
act as {\em auxiliary} from the view of function $f$. We note that the model
counts of $f$ and $g$ are not necessarily the same.

We utilize this notion in compiling DNNFs as shown in Algorithm~\ref{alg:dnnf}.
Here, to compile a DNNF representation of a function $f(\X)$, we first obtain
another function $g(\X,\Y)$ that is emf to function $f$, with variables $\Y$
being auxiliary (Line~\ref{ln:emf}). Clearly, the specific method to construct
function $g$ would depend on the input representation of $f$. We will discuss
different ways for that later in Section~\ref{sec:emf_trans} when the input is a
CNF. Once function $g$ is constructed, we compile a deterministic DNNF
representation of it using an off-the-shelf knowledge compiler
(Line~\ref{ln:compile}). Finally, we forget auxiliary variables $\Y$ from the
compiled structure (Line~\ref{ln:forget}). This would generate a DNNF
representation of the input as $g$ is emf to function $f$.

\begin{proposition}
Algorithm~\ref{alg:dnnf} returns a DNNF representation of its input.
\end{proposition}

\begin{algorithm}[t]
\LinesNumbered
\DontPrintSemicolon
\caption{$DNNF(f)$}
\label{alg:dnnf}
\KwIn{$f(\X):$ a Boolean function over variables $\X$}
\KwOut{constructs a DNNF representation of function $f$}

\BlankLine\hrule\BlankLine

$g(\X,\Y) \lar emf(f)$\label{ln:emf}\;
$\Delta \lar$ compile $g(\X,\Y)$ using a d-DNNF compiler\label{ln:compile}\;
$\Gamma \lar$ forget variables $\Y$ from $\Delta$\label{ln:forget}\;
\KwRet{$\Gamma$}

\end{algorithm}

\begin{figure}[t]
  \centering
  \subfigure[Before forgetting]{
    \begin{tikzpicture}[->, >=stealth', line width=1pt,
                   level 1/.style={sibling distance = 2.5cm, level distance = 1.25cm},
                   level 2/.style={sibling distance = 1.5cm, level distance = 1.25cm},
                   level 3/.style={sibling distance = 1.5cm, level distance = 1.25cm},
                   level 4/.style={sibling distance = 1.25cm}]

\tikzstyle{nnf} = [minimum size=.25cm]
\tikzset{every node/.style={nnf}}

\node {$\vee$}
  child{ 
    node {$\wedge$}
      child{node {$X$}}
      child{
        node {$\vee$}
          child{
            node {$Y$}
          }
          child{
            node {$\wedge$}
              child{node {$\neg Y$}}
              child{node (z) {$Z$}}
              child{node (q) {$\neg Q$}}
          }
      }
  }
  child{
    node {$\wedge$}
      child{node {$\neg X$}}
      child{
        node (a) {$\wedge$} 
      }
  };

\draw [->] (a) -- (z);
\draw [->] (a.south) -- (q);
\end{tikzpicture}
  }\qquad
  \subfigure[After forgetting]{
    \begin{tikzpicture}[->, >=stealth', line width=1pt,
                   level 1/.style={sibling distance = 2.5cm, level distance = 1.25cm},
                   level 2/.style={sibling distance = 1.5cm, level distance = 1.25cm},
                   level 3/.style={sibling distance = 1.5cm, level distance = 1.25cm},
                   level 4/.style={sibling distance = 1.25cm}]

\tikzstyle{nnf} = [minimum size=.25cm]
\tikzset{every node/.style={nnf}}

\node {$\vee$}
  child{ 
    node {$\wedge$}
      child{node {$\top$}}
      child{
        node {$\vee$}
          child{
            node {$Y$}
          }
          child{
            node {$\wedge$}
              child{node {$\neg Y$}}
              child{node (z) {$\top$}}
              child{node (q) {$\neg Q$}}
          }
      }
  }
  child{
    node {$\wedge$}
      child{node {$\top$}}
      child{
        node (a) {$\wedge$} 
      }
  };

\draw [->] (a) -- (z);
\draw [->] (a.south) -- (q);
\end{tikzpicture}
  }
  \caption{Forgetting variables $X,Z$ from a DNNF.}
  \label{fig:forget}
\end{figure}
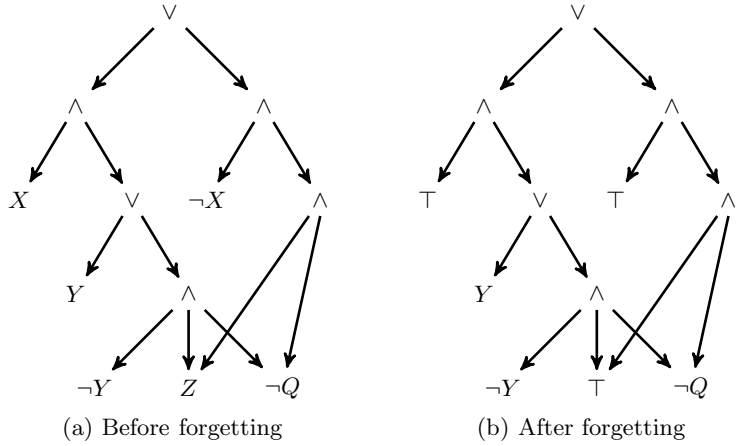

We remark that the last step of Algorithm~\ref{alg:dnnf} can be performed only
in linear time in the size of the structure. This is due to the property of
decomposability, which supports linear time multiple-variable forgetting: all
one needs is to replace auxiliary variables with the constant $\top$ in the
structure. An example of this procedure is depicted in Fig.~\ref{fig:forget},
where we forget variables $X,Z$ from a deterministic DNNF. What is crucial
here is that the resulting structure does not enforce determinism anymore, but
the decomposability property stays intact. In fact, as we will show in the next
section, this could lead to exponentially more succinct representations, which
can be thought of as a compensation for losing the ability of performing
efficient model counting.

\section{An Exponential Separation by Forgetting Variables}\label{sec:forget}

In this section, we address the following question: to what extend forgetting
auxiliary variables could lead to more compact DNNFs without determinism than
deterministic DNNFs? 

We next state our main result, showing that exponentially more compact
representations can be obtained.
\begin{theorem}\label{thm:sep}
There exist two classes of Boolean functions $f_n(\X)$ and $g_n(\X,Z)$ such
that: (i) $f_n$ is emf to $g_n$, (ii) the size of each d-DNNF computing $f_n$ is
at least exponential in $n$, and (iii) there is a d-DNNF computing $g_n$ whose
size is polynomial in $n$.
\end{theorem}

\noindent In other words, it is not feasible to compile a deterministic DNNF
representation of $f_n$, yet one can construct a compact DNNF representation of
$f_n$ through forgetting a single auxiliary variable from the compact
deterministic DNNF representation of $g_n$. We remark that obtaining a DNNF
computing $f_n$ directly is not possible in practice as existing knowledge
compilers generate deterministic subsets of DNNF. 

We next present the proof of Theorem~\ref{thm:sep}, where we make use of the
function that has been shown to exponentially separate DNNFs from deterministic
DNNFs~\cite{BovaCMS16}.

Let $\M$ be an $n \times n$ matrix of Boolean variables. Let $\R_1, \ldots,
\R_n$ be the rows of $M$ and $\C_1, \ldots, \C_n$ be the columns of $M$. Let
$h_n$ be the class of functions over $n$ variables evaluating to 1 iff the sum
of its inputs is divisible by 3. Consider the following function defined on the
variables of $\M$ and variable $Z$: 
$$
g_n(\M,Z) = \big(Z \wedge row_n(\M)\big) \vee \big(\neg Z \wedge col_n(\M)\big),
$$
where $row_n$ and $col_N$ are defined by
$$
row_n(\M) = \bigoplus\limits_{i=1}^{n} h_n(\R_i),\qquad
col_n(\M) = \bigoplus\limits_{i=1}^{n} h_n(\C_i).
$$

\noindent Finally, let $f_n$ be the following function defined on the variables
of $\M$:
$$
f_n(\M) = row_n(\M) \vee col_n(\M).
$$

\noindent Clearly, $f_n(\M) \equiv \exists\,Z.\, g_n(\M,Z)$, and hence function
$f_n$ is emf to function $g_n$. Indeed, function $f_n$ is the {\em Sauerhoff}
function~\cite{Sauerhoff03}, which was used in the exponential separation of
DNNFs from deterministic DNNFs~\cite{BovaCMS16}. That is, $f_n$ has a polynomial
size DNNF representation, but each deterministic DNNF computing it is
exponential in size. Finally, since functions $row_n$ and $col_n$ both have
polynomial size OBDDs (a subset of deterministic DNNF), function $g_n$ has a
polynomial size deterministic DNNF representation. Thus, Theorem~\ref{thm:sep}
holds. 

As a side note, this result implies that forgetting on deterministic DNNF and
FBDD cannot be done in polynomial time, which was only known up to some standard
complexity-theoretic assumptions (i.e., $\P \neq \NP$).\footnote{The same
function can also be used to show that d-DNNF and FBDD do not support
polynomial time disjunction operation.}

\begin{corollary} 
d-DNNF and FBDD do not support polynomial time single-variable forgetting, as
well as polynomial time multiple-variable forgetting.
\end{corollary}

Theorem~\ref{thm:sep} reveals the usefulness of our new approach in theory. To
make it useful in practice, we need to identify transformations that would
produce emf formulas, which is discussed next.

\section{EMF Transformations}\label{sec:emf_trans}

In this section, we address the following question: how can one identify
functions that are equivalent modulo forgetting?

We will study some existing techniques for CNFs that incorporate auxiliary
variables, mostly to get an equisatisfiable CNF. For each technique, we will
demonstrate that the produced equisatisfiable CNF is indeed emf to the input
CNF. We first formally define a notion of transformation that will be used to
identify methods producing emf formulas.

\begin{definition}
Let $T$ be an algorithm that takes as input a Boolean function $f(\X)$ and
outputs another Boolean function $g(\X,\Y)$, where $\X$ and $\Y$ are disjoint.
Then algorithm $T$ is said to be an \underline{emf transformation} iff function
$f$ is emf to function $g$.
\end{definition}

\noindent Given this definition, we next present some emf transformations that
exist in the literature. 

\subsection{Tseitin Transformation}

State-of-the-art SAT solvers require their input to be a Boolean formula in CNF.
When this is not the case, one has to first transform the input into a CNF. The
naive approach here is to use the famous De Morgan's law and the distributive
property, which preserves logical equivalence. However, this can easily blow-up
CNF size exponentially. Thus, one typically applies Tseitin
transformation~\cite{Tseitin68}, which converts a Boolean formula into an
equisatisfiable CNF by adding auxiliary variables with only a linear increase in
size. In fact, Tseitin transformation does more than constructing an
equisatisfiable CNF. In particular, it guarantees two more
properties~\cite{Tseitin68}:

\begin{itemize}

  \item[(1)] Dropping auxiliary variables from a model of the constructed CNF
    would yield a model of the input formula;
    
  \item[(2)] Any model of the input formula can be extended to be a model of the
    constructed CNF.

\end{itemize}

\noindent As we prove next, these two properties make Tseitin transformation an
emf transformation, as well as any other transformation that satisfies them.
\begin{theorem}\label{thm:tseitin}
Let $T$ be a transformation that satisfies the two properties above. Then $T$ is
an emf transformation.
\end{theorem}

Let $f(\X)$ be the input function to transformation $T$, and let $g(\X,\Y)$ be
the function constructed for $f(\X)$ by transformation $T$, where variables $\Y$
are introduced during the transformation. We want to show that $f(\X) \equiv
\exists\,\Y.\, g(\X,\Y)$.

Let $\x$ be a model of $\exists\,\Y.\, g(\X,\Y)$. We will show that $\x$ is also
a model of $f(\X)$. Since $\x$ is a model of $\exists\,\Y.\, g(\X,\Y)$, there
must be an instantiation $\y$ such that $\x\y$ is a model of $g(\X,\Y)$. Then,
by the first property above, $\x$ must be a model of $f(\X)$.

Let $\x$ be a model of $f(\X)$. We will show that $\x$ is also a model of
$\exists\,\Y.\, g(\X,\Y)$. Due to the second property above, there must exist an
instantiation $\y$ such that $\x\y$ is a model of $g(\X,\Y)$.  Then, as
$\exists\,\Y.\, g(\X,\Y)$ says everything $g(\X,\Y)$ says on variables $\X$,
$\x$ must be a model of $\exists\,\Y.\, g(\X,\Y)$.

Therefore, Theorem~\ref{thm:tseitin} holds, which immediately implies that
Tseitin transformation is an emf transformation.

\begin{proposition}
Tseitin transformation is an emf transformation.
\end{proposition}

\noindent Accordingly, we can apply Tseitin transformation to compile DNNF when
the input is not in CNF, which is also the required format for most knowledge
compilers.

\subsection{Extended Resolution}

Resolution is a powerful rule of inference that has been used in SAT
solving~\cite{Robinson65}. Specifically, iterating the following rule repeatedly
in a certain way would tell whether a CNF is satisfiable or not: 

$$
\frac{X \vee \alpha \quad \neg X \vee \beta}{\alpha \vee \beta},
$$

\noindent where $X$ is a variable and $\alpha$ and $\beta$ are clauses. This
rule states that whenever the clauses in the premise appear in a CNF, one can
increment the CNF by adding the clause in the conclusion, without changing the
logical content of the CNF (i.e., preserving logical equivalence). Here, $\alpha
\vee \beta$ is called the {\em resolvent} obtained by {\em resolving} variable
$X$ on $X \vee \alpha$ and $\neg X \vee \beta$.

It turns out that resolution could generate only exponentially long proofs of
unsatisfiability for certain families of formulas (see, e.g., the Pigeonhole
principle~\cite{Haken85}). To remedy this, {\em extended} resolution is
introduced, which is a more powerful generalization of resolution that includes
an additional rule, called the {\em extension} rule~\cite{Tseitin68}.
Accordingly, extended resolution allows one to increment the CNF with the
addition of clauses of the form $X \iff \ell_1 \vee \ell_2$\footnote{More
specifically, $X \iff \ell_1 \vee \ell_2$ can be replaced with the clauses $\neg
X \vee \ell_1 \vee \ell_2,\, X \vee \neg \ell_1,\, X \vee \neg \ell_2$.}, where
$X$ is an auxiliary variable that does not appear in the CNF and literals
$\ell_1$ and $\ell_2$ appear in the CNF. Then one can apply the resolution rule
as before. This simple addition creates an exponentially more powerful proof
system than resolution, as extended resolution could generate polynomial size
proofs where the regular resolution can only generate exponential size
proofs~\cite{Cook76}.

Indeed, extended resolution constructs an equisatisfiable CNF, and thus applying
resolution on it produces correct results for SAT solving. This technique has
also been shown to be useful in practice of SAT solving, where different schemes
for applying the extension rule have been
suggested~\cite{Huang10,AudemardKS10,Manthey14}.  Hence, its usage could
potentially be extended to DNNF compilation, given that we will now show it is
indeed an emf transformation.

We will now prove the following result, which generalizes extended resolution.
\begin{theorem}
Let $f(\X),\alpha^1(\X),\ldots,\alpha^n(\X)$ be Boolean functions. Consider the
class of Boolean functions
$$
g_n(\X,\Y) =
f(\X) \wedge (Y_1 \iff \alpha^1(\X)) \wedge \ldots \wedge (Y_n \iff \alpha^n(\X)),
$$ 
where $\Y = \{Y_1,\ldots,Y_n\}$. Then function $f$ is emf to function $g_n$.
\end{theorem}

We want to show that $f(\X) \equiv \exists\,\Y.\, g_n(\X,\Y)$. For that, we will
use the following simplification $n$ times:
\begin{align}
\exists\,\Y.\, g_n
 & \equiv \exists\,Y_1,\ldots,Y_{n-1}.\,\, 
     \exists\,Y_n.\, f(\X) \wedge \bigwedge\limits_{i=1}^{n} Y_i \iff
     \alpha^i(\X)\label{eq:1}\\
 & \equiv \exists\,Y_1,\ldots,Y_{n-1}.\,\,
     f(\X) \wedge \bigg(\bigwedge\limits_{i=1}^{n-1} Y_i \iff \alpha^i(\X)\bigg)
     \wedge\, \exists\,Y_n.\, Y_n \iff \alpha^n(\X)\label{eq:2}\\
 & \equiv \exists\,Y_1,\ldots,Y_{n-1}.\, 
   f(\X) \wedge \bigwedge\limits_{i=1}^{n-1} Y_i \iff \alpha^i(\X)\label{eq:3}\\
 & \cdots\nonumber \\
 & \equiv \exists\,Y_1.\, f(\X) \wedge \big(Y_1 \iff \alpha_1(\X)\big)\nonumber \\ 
 & \equiv f(\X).\nonumber
\end{align}

\noindent Equation~(\ref{eq:1}) is due to the definition of multiple-variable
forgetting. Equation~(\ref{eq:2}) holds as $f(\X) \wedge
\bigwedge\limits_{i=1}^{n-1} Y_i \iff \alpha_i(\X)$ does mention variable $Y_n$.
Equation~(\ref{eq:3}) holds as forgetting variable $Y_n$ from $Y_n \iff
\alpha_n(\X)$ is equivalent to the trivial function~$\top$. 

Assuming that $f(\X)$ is a CNF, replacing each $\alpha^i(\X)$ with a clause of
two literals of variables $\X$ would clearly correspond to the extension rule of
extended resolution.

\begin{proposition}
Extended resolution is an emf transformation.
\end{proposition}

\subsection{Bounded Variable Addition}

Bounded variable addition (BVA) is a preprocessing technique introduced for SAT
solving~\cite{MantheyHB12}. The goal here is to reduce the sum of the number of
variables and clauses of a CNF by introducing auxiliary variables, without
losing the ability of answering the SAT query. It is based on resolution as
described next.

Let $C_X$ be a set of clauses containing literal $X$ and $C_{\neg X}$ a set of
clauses containing literal $\neg X$. Let $C_X \bowtie C_{\neg X}$ denote the set
of resolvents one would obtain by resolving $X$ on clauses in $C_X$ and $C_{\neg
X}$. Given a CNF $\Delta$ and an auxiliary variable $X$ that does not appear in
$\Delta$, BVA looks for sets of clauses $C_X$ and $C_{\neg X}$ such that $C_X
\bowtie C_{\neg X}$ belongs to $\Delta$ and $|C_X \bowtie C_{\neg X}| > |C_X| +
|C_{\neg X}|$. In this case, BVA replaces clauses $C_X \bowtie C_{\neg X}$ with
clauses $C_X$ and $C_{\neg X}$. For instance, consider the following CNF:
$$
\Delta = (A \vee D) \wedge (B \vee D) \wedge (C \vee D) \wedge 
         (A \vee E) \wedge (B \vee E) \wedge (C \vee E).
$$

\noindent By adding an auxiliary variable $X$, we can obtain the following CNF
which has fewer clauses than $\Delta$:
$$
\Sigma = (A \vee \neg X) \wedge (B \vee \neg X) \wedge  (C \vee \neg X) \wedge
         (D \vee X) \wedge (E \vee X). 
$$

\noindent Indeed, $\Delta$ is equisatisfiable to $\Sigma$, and thus one can feed
$\Sigma$ to a SAT solver, instead of $\Delta$. 

The authors of~\cite{MantheyHB12} also developed a heuristic to apply the BVA
transformation on CNFs, which is a greedy algorithm that searches for
clause-patterns in the input CNF. This algorithm is shown to be useful in SAT
solving, and thus offering a practical idea to apply to our DNNF compilation
method due to the following result. 

\begin{proposition}
Bounded variable addition is an emf transformation.
\end{proposition}

We will now prove the above proposition. In particular, let $\Delta(\X)$ be a
CNF and $\Sigma(\X,Y)$ be the CNF obtained by applying BVA on $\Delta$, where
$Y$ is the auxiliary variable added during the process. Then we want to show
that $\Delta(\X) \equiv \exists\,Y.\, \Sigma(\X,Y)$.

Let $\Gamma_Y = \wedge_{i=1}^m Y \vee \alpha_i$ and $\Gamma_{\neg Y} =
\wedge_{j=1}^k \neg Y \vee \beta_j$ be the clauses containing literals $Y$ and
$\neg Y$ in CNF $\Sigma$, respectively. Then, due to the BVA process, we can
rewrite CNF $\Delta$ as the CNF $\Phi \wedge \Gamma_Y \bowtie \Gamma_{\neg Y}$
where $\Phi$ is a CNF. Moreover, CNF $\Sigma$ is equivalent to $\Phi \wedge
\Gamma_Y \wedge \Gamma_{\neg Y}$. In this setting, we have the following
equations:
\begin{align*}
\exists\,Y.\, \Sigma(\X,Y) 
  & \equiv \exists\,Y.\, \Phi \wedge \Gamma_Y \wedge \Gamma_{\neg Y} \\
  & \equiv \Phi \wedge \exists\,Y.\, \Gamma_Y \wedge \Gamma_{\neg Y} \\
  & \equiv \Phi \wedge \Big( 
                             \big(\Gamma_Y \wedge \Gamma_{\neg Y}\big)|Y \vee
                             \big(\Gamma_Y \wedge \Gamma_{\neg Y}\big)|\neg Y
                       \Big) \\
  & \equiv \Phi \wedge \Big( \Gamma_{\neg Y}|Y \vee \Gamma_Y|\neg Y \Big) \\
  & \equiv \Phi \wedge \Big( 
                             \bigwedge\limits_{j=1}^k \beta_j \vee
                             \bigwedge\limits_{i=1}^m \alpha_i
                       \Big) \\
  & \equiv \Phi \wedge \Big( 
                             \bigwedge\limits_{j=1}^k
                             \bigwedge\limits_{i=1}^m \beta_j \vee \alpha_i
                       \Big) \\
  & \equiv \Phi \wedge \Gamma_Y \bowtie \Gamma_{\neg Y}\\ 
  & \equiv \Delta(\X).
\end{align*}

\subsubsection{Treewidth and BVA} We now identify another guarantee that comes
with the BVA transformation. Treewidth\footnote{The definition of treewidth and
some of its properties is delegated to Appendix~\ref{app:width}.} is a
well-known graph-theoretic property~\cite{RobertsonS84}, which has been
extensively used as a parameter that renders many hard reasoning tasks
tractable when being small. In the context of knowledge compilation, it is
known that compiling a CNF into a deterministic DNNF can be done in the worst
case in time that is linear in the number of variables and exponential in the
treewidth of the CNF primal graph~\cite{Darwiche01}.\footnote{Primal graph is
a CNF abstraction that represents the connection between the variables and
clauses of the CNF.  In particular, each vertex of the graph represents a
variable, and there is an edge between to vertices iff the corresponding
variables appear together in one of the CNF clauses.} Therefore, a CNF with a
bounded treewidth can easily be compiled into a deterministic DNNF. 

We will next present two results regarding the effects of the BVA transformation
on the primal treewidth of CNFs, whose proofs are delegated to
Appendix~\ref{app:width}. Our first result is the following guarantee.
\begin{theorem}\label{thm:width}
Let $\Delta$ be a CNF whose primal treewidth is $w$. Let $\Sigma$ be the CNF
obtained by applying the BVA transformation $k$ times on CNF $\Delta$. Then the
primal treewidth of $\Sigma$ is at most $w+k$.
\end{theorem}

\noindent Hence, the BVA transformation would not affect the treewidth much in
the worst case, when applied constant times. Moreover, as we present next, the
BVA transformation could potentially reduce the treewidth from an unbounded
value to a bounded value. 
\begin{theorem}\label{thm:bva}
There exists a class of CNFs $\Delta_n$ over $n^3$ variables such that: (i) the
primal treewidth of $\Delta_n$ is unbounded (i.e., at least $n$), and (ii)
applying the BVA transformation 2 times on $\Delta_n$ can generate a CNF whose
primal treewidth is bounded (i.e., at most 2). 
\end{theorem}

\noindent Theorem~\ref{thm:bva} implies that the BVA transformation can generate
a CNF whose compilation to deterministic DNNF is easy, whereas this cannot be
identified in the input CNF (as the treewidth is unbounded). Therefore,
Algorithm~\ref{alg:dnnf} can easily compile a DNNF in this case, if the BVA
transformation is applied. Yet, there is no guarantee on compiling a
deterministic DNNF with existing compilers, without applying the BVA
transformation. Indeed, we will confirm this empirically in our experiments,
where the following class of CNFs $\Delta_n^a$ will be considered:
$$
\bigwedge\limits_{1 \leq i,j,k \leq n} X_i \vee Y_j \vee Z_k.
$$

\noindent This class of CNFs has unbounded treewidth. On the other hand, the
following class of CNFs $\Delta_n^b$ can be identified by the BVA
transformation, which has bounded treewidth.
$$
\Big(\bigwedge\limits_{1 \leq i \leq n} A \vee X_i\Big)  \wedge
\Big(\bigwedge\limits_{1 \leq j \leq n} \neg A \vee B \vee Y_j\Big)\wedge
\Big(\bigwedge\limits_{1 \leq k \leq n} \neg B \vee Z_k\Big).
$$

\noindent Note that we added two auxiliary variables $A,B$ into CNF
$\Delta_n^a$, and reduced the number of clauses from $n^3$ to $3n$.

\section{Experiments}\label{sec:exp}

\begin{table}[t]
\centering
\def\arraystretch{1.5}
\setlength\tabcolsep{5pt}
\caption{Experimental results on CNFs $\Delta_n^a$. \cdf is our approach,
compiling DNNFs without determinism. All timings are in seconds.}
\label{tab:exp1}
\begin{tabular}{c||rrr|rrr}
  & \multicolumn{3}{c|}{\cdf}& \multicolumn{3}{c}{\cd} \\
$\Delta_n^a$ & \#node  & \#edge  & Time    & \#node & \#edge & Time \\
\hline
10  & 42  & 43  & 0.04 & 794   & 1,578  & 0.11 \\
15  & 57  & 58  & 0.03 & 26,199 & 52,368 & 11.43 \\
30  & 102 & 103 & 0.04 & --     & --     & --  \\
50  & 162 & 163 & 0.04 & --     & --     & --  \\
75  & 237 & 238 & 0.04 & --     & --     & --  \\
100 & 312 & 313 & 0.04 & --     & --     & --  \\
\end{tabular}
\end{table}

In this section, we will empirically demonstrate the applicability of
Algorithm~\ref{alg:dnnf} in compiling DNNFs, when coupled with the BVA
transformation. In particular, we compile CNFs into DNNF and deterministic DNNF.
For the latter we use the \cd compiler\footnote{Available at
\texttt{http://reasoning.cs.ucla.edu/c2d.}}, and for the former we use the
same compiler after preprocessing CNFs by the preprocessor
\coproc\footnote{Available at
\texttt{http://tools.computational-logic.org/content/riss.php}.} and
forgetting auxiliary variables after the compilation.

We evaluated the mentioned systems on two different benchmarks. First, we used
the manually constructed class of CNFs $\Delta_n^a$ (described in
Section~\ref{sec:emf_trans}) for values of $n \in \{10, 15, 30, 50, 75, 100\}$.
Second, we used some CNF encodings of wire routing problems in the channels of
field-programmable gate arrays (FPGA)~\cite{NamASR04}. The goal here is to
decide if a routing configuration is possible. That is, given $m$ connections
and $k$ channels on an FPGA (denoted fpga\_m\_n), the satisfiability of the CNF
encoding would imply that the routing of $m$ connections through $k$ channels is
possible. Our experiments were performed on a 2.6GHz Intel Xeon E5-2670 CPU with
a 1 hour time limit and a memory limit of 8GB RAM.

Table~\ref{tab:exp1} highlights the results on CNFs $\Delta_n^a$. According to
this, our approach (\cdf) recognizes the tractability of the CNF instances by
introducing two auxiliary variables (as shown in Section~\ref{sec:emf_trans}),
and thus it compiles the instances quickly and compactly. On the other hand, the
traditional approach (\cd) performed poorly as it could not finish compilation
after $n=20$.

For the FPGA routing problems, we first present some statistics of the CNF
instances before and after the preprocessing in Table~\ref{tab:stats}. We now
highlight the results in Table~\ref{tab:exp2}. Accordingly, our approach is
clearly superior than the traditional approach as we can compile 5 instances
which otherwise could not be compiled. In the remaining 2 instances, not only
our approach produces DNNFs faster but also constructs more compact
representations. Therefore, our approach improves performance of DNNF
compilation on these FPGA problems.

\begin{table}[t]
\centering
\def\arraystretch{1.5}
\setlength\tabcolsep{5pt}
\caption{Some stats on CNF encodings of FPGA routing problems, before and after
preprocessing.}
\label{tab:stats}
\begin{tabular}{c||rr|rrc}
  & \multicolumn{2}{c|}{Before BVA}& \multicolumn{3}{c}{After BVA} \\
Instance   & \#variable  & \#clause   & \#variable & \#clause & \#aux\_variable\\
\hline
fpga\_10\_8  & 120 & 448  & 158 & 290 & 38 \\
fpga\_10\_9  & 135 & 549  & 174 & 330 & 39 \\
fpga\_12\_8  & 144 & 560  & 188 & 356 & 44 \\
fpga\_12\_9  & 162 & 684  & 207 & 405 & 45 \\
fpga\_12\_11 & 198 & 968  & 269 & 503 & 71 \\
fpga\_12\_12 & 216 & 1128 & 300 & 552 & 84 \\
fpga\_13\_9  & 176 & 759  & 229 & 444 & 53 \\
\end{tabular}
\end{table}

\section{Related Work}\label{sec:related}

The closest related work to ours is perhaps the work of~\cite{PipatsrisawatD08},
in which the authors identified a subset of DNNF, called {\em structured} DNNF.
The significance here is that this subset supports a polynomial time conjoin
operation~\cite{PipatsrisawatD08}, while general DNNF do not support this
(unless $\P = \NP$)~\cite{DarwicheM02}. Due to this operation, one can compile
CNFs incrementally in a bottom-up fashion into a structured DNNF. That is, after
representing each clause as a structured DNNF (which can be done easily), one
can conjoin clauses one by one until a structured DNNF is compiled for the input
CNF. Indeed, the compiled DNNF would not necessarily be deterministic (as the
conjoin operation does not enforce this). However, building an efficient
knowledge compiler based on this approach would require intensive engineering
effort and has not been accomplished yet. Our work, on the other hand, leverages
state-of-the-art knowledge compilers as it only depends on constructing emf
formulas. Due to this, one can quickly build an efficient DNNF compiler, as we
have done in this work. Moreover, DNNF could be exponentially more succinct than
its structured subset, which could make the mentioned work more restrictive than
our presented approach. 

\begin{table}[t]
\centering
\def\arraystretch{1.5}
\setlength\tabcolsep{5pt}
\caption{Experimental results on FPGA routing problems. \cdf is our approach,
compiling DNNFs without determinism. All timings are in seconds.}
\label{tab:exp2}
\begin{tabular}{c||rrr|rrr}
  & \multicolumn{3}{c|}{\cdf}& \multicolumn{3}{c}{\cd} \\
Instance   & \#node  & \#edge  & Time    & \#node & \#edge & Time \\
\hline
fpga\_10\_8  & 38,601   & 116,399  & 0.66    & 122,106 & 398,915 & 97.63  \\
fpga\_10\_9  & 37,528   & 107,316  & 0.84    & 199,563 & 695,470 & 661.85  \\
fpga\_12\_8  & 215,790  & 595,522  & 26.87   & --     & --     & --  \\
fpga\_12\_9  & 428,340  & 1,303,189 & 92.68   & --     & --     & --  \\
fpga\_12\_11 & 491,225  & 1,428,101 & 99.43   & --     & --     & --  \\
fpga\_12\_12 & 389,274  & 1,115,493 & 207.58  & --     & --     & --  \\
fpga\_13\_9  & 1,149,770 & 3,133,399 & 268.34  & --     & --     & --  \\
\end{tabular}
\end{table}

Another related work to ours is that of~\cite{LagniezM14,LagniezM17}, in which
the authors studied the effects of preprocessing CNFs for model counting. They
considered various techniques from the literature and also introduced a few new
ones, which resulted in an efficient preprocessor. Their focus was on
constructing CNFs that are either equivalent or preserving the model count. Our
work is based on preprocessing techniques that generate emf formulas, and
targets compiling DNNFs, as opposed to counting the models. 

Finally, \cite{AzizCMS15} studied the problem of {\em projected} model counting,
in which the goal is to compute the model count of a formula after forgetting
certain variables. In their setting, auxiliary variables are named as
``non-priority'' variables. The main distinction here is that we are not
interested in the model counting after forgetting variables. Because of this,
interestingly enough, the forgetting operation helps in our setting to obtain
more compact representations.

\section{Conclusion}\label{sec:conc}

In this work, we studied compiling DNNFs without enforcing determinism. We
presented a new methodology to relax determinism, which is based on introducing
auxiliary variables and forgetting them from a deterministic DNNF. We
demonstrated that several existing techniques that introduce auxiliary variables
can be used in our framework, allowing us to exploit existing knowledge
compilers. We further showed that our new approach can lead to exponentially
more compact representations, and our experimental evaluation confirmed the
applicability of the new technique on certain benchmarks, when bounded variable
addition is employed to introduce auxiliary variables.

\appendix

\section{Treewidth}\label{app:width}

In this section, we will define treewidth and present the proofs of
Theorem~\ref{thm:width} and Theorem~\ref{thm:bva}.  We start with a definition
of the primal treewidth of a CNF, where we choose to use the one based on {\em
jointrees} (e.g., \cite{Darwiche09}) among a number of ways.

A jointree for a CNF $\Delta$ is a tree whose vertices are labeled with a subset
of variables of $\Delta$ such that the following two conditions hold:
\begin{itemize}

  \item For each clause $\gamma$ of $\Delta$, there is a vertex whose labels
    contain the variables of $\gamma$; 

  \item If a variable $X$ appears in the labels of two vertices $V_1$ and $V_2$,
    then each vertex on the path connecting $V_1$ and $V_2$ includes variable
    $X$ in its labels. 

\end{itemize}

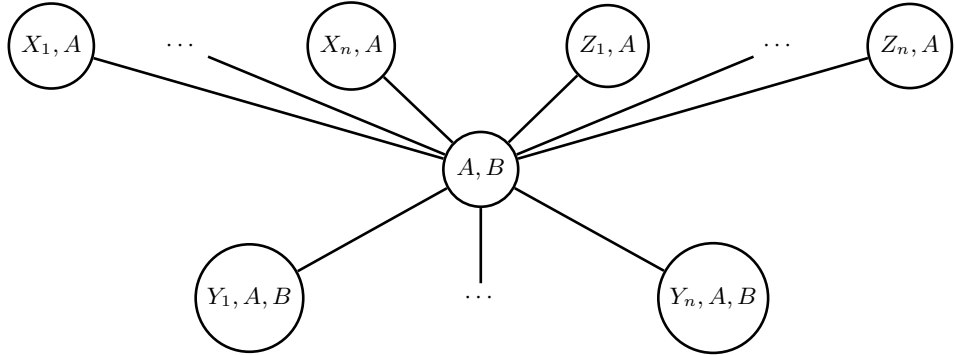
\begin{figure}[t]
  \centering
  \begin{tikzpicture}[line width=1pt]

\tikzstyle{vertex} = [draw, circle, minimum size=.5cm]

\node[vertex] (root) {$A, B$};

\node [above = 1cm of root] (xz) {};

\node [vertex, left = 5cm of xz] (x1) {$X_1, A$};
\node [left = 3.5cm of xz] (xmid) {$\cdots$};
\node [vertex, left = 1cm of xz] (x2) {$X_n, A$};

\node [vertex, right = 1cm of xz] (z1) {$Z_1, A$};
\node [right = 3.5cm of xz] (zmid) {$\cdots$};
\node [vertex, right = 5cm of xz] (z2) {$Z_n, A$};

\node [below = 1cm of root] (ymid) {$\cdots$};
\node [vertex, left= 2cm of ymid]  (y1) {$Y_1, A, B$};
\node [vertex, right= 2cm of ymid] (y2) {$Y_n, A, B$};

\draw (x1) -- (root);
\draw (x2) -- (root);
\draw (xmid) -- (root);

\draw (z1) -- (root);
\draw (z2) -- (root);
\draw (zmid) -- (root);

\draw (y1) -- (root);
\draw (y2) -- (root);
\draw (ymid) -- (root);

\end{tikzpicture}
  \caption{A jointree for CNF $\Delta_n^b$.}
  \label{fig:jointree}
\end{figure}

\noindent The labels of a vertex of a jointree is called its {\em cluster}. The
{\em width} of a jointree is the size of its largest cluster minus 1. The primal
treewidth of a CNF is the smallest width attained by any of its jointrees.  For
instance, Fig.~\ref{fig:jointree} depicts a jointree for $\Delta_n^b$ whose
width is 2.

\spara{Proof of Theorem~\ref{thm:width}} Assume that we apply the BVA
transformation on CNF $\Delta$ once, and constructed the CNF $\Delta^1$. So, an
auxiliary variable $X$ is added to CNF $\Delta^1$. Consider now the best
jointree of $\Delta$ (i.e., the one whose width is $w$). If we add variable $X$
to each label set of its vertices, the resulting tree will clearly be a jointree
for $\Delta^1$, with width $w+1$. So, the treewidth of $\Delta^1$ will be at
most $w+1$. Now, if we apply the same idea after each application of the BVA
transformation, the treewidth will be at most $w+k$ after the $k^{th}$ step.

\spara{Proof of Theorem~\ref{thm:bva}} We first show that the treewidth of
$\Delta_n^a$ is unbounded (i.e., at least $2n$). In the primal graph of
$\Delta_n^a$, each vertex will have a degree of $2n$. According to a known
result (see, e.g.,~\cite{Darwiche09}), this implies that the treewidth of
$\Delta_n^a$ is no less than $2n$.

We will now show that the treewidth of $\Delta_n^b$, which can be obtained from
$\Delta_n^a$ by the BVA transformation, is bounded (i.e., at most 2).
Figure~\ref{fig:jointree} depicts a jointree for $\Delta_n^b$ whose width is 2.
Hence, the treewidth of $\Delta_n^b$ is at most 2.

\bibliography{references}

\begin{thebibliography}{10}
\providecommand{\url}[1]{\texttt{#1}}
\providecommand{\urlprefix}{URL }

\bibitem{AudemardKS10}
Audemard, G., Katsirelos, G., Simon, L.: {A Restriction of Extended Resolution
  for Clause Learning Sat Solvers}. In: Proceedings of the Twenty-Fourth AAAI
  Conference on Artificial Intelligence. pp. 15--20 (2010)

\bibitem{AzizCMS15}
Aziz, R.A., Chu, G., Muise, C., Stuckey, P.: {$\#\exists$ SAT: Projected Model
  Counting}. In: Proceedings of the Eighteenth International Conference on
  Theory and Applications of Satisfiability Testing. pp. 121--137 (2015)

\bibitem{Barrett05}
Barrett, A.: {Model Compilation for Real-Time Planning and Diagnosis with
  Feedback}. In: Proceedings of the Nineteenth International Joint Conference
  on Artificial Intelligence. pp. 1195--1200 (2005)

\bibitem{BovaCMS16}
Bova, S., Capelli, F., Mengel, S., Slivovsky, F.: {Knowledge Compilation Meets
  Communication Complexity}. In: Proceedings of the Twenty-Fifth International
  Joint Conference on Artificial Intelligence. pp. 1008--1014 (2016)

\bibitem{Bryant86}
Bryant, R.E.: {Graph-Based Algorithms for Boolean Function Manipulation}. IEEE
  Transactions on Computers  35(8),  677--691 (1986)

\bibitem{ChaviraD08}
Chavira, M., Darwiche, A.: {On Probabilistic Inference by Weighted Model
  Counting}. Artifical Intelligence  172(6-7),  772--799 (2008)

\bibitem{Cook76}
Cook, S.A.: {A Short Proof of the Pigeon Hole Principle Using Extended
  Resolution}. SIGACT News  8(4),  28--32 (1976)

\bibitem{Darwiche00}
Darwiche, A.: {Model-Based Diagnosis under Real-World Constraints}. AI Magazine
   21(2),  57--73 (2000)

\bibitem{Darwiche01}
Darwiche, A.: {Decomposable Negation Normal Form}. Journal of the ACM  48(4),
  608--647 (2001)

\bibitem{Darwiche00b}
Darwiche, A.: {On the Tractable Counting of Theory Models and its Application
  to Truth Maintenance and Belief Revision}. Journal of Applied Non-Classical
  Logics  11(1-2),  11--34 (2001)

\bibitem{Darwiche02b}
Darwiche, A.: {A Logical Approach to Factoring Belief Networks}. In:
  Proceedings of the Eighth International Conference on Principles of Knowledge
  Representation and Reasoning. pp. 409--420 (2002)

\bibitem{Darwiche04}
Darwiche, A.: {New Advances in Compiling CNF into Decomposable Negation Normal
  Form}. In: Proceedings of the Sixteenth European Conference on Artificial
  Intelligence. pp. 328--332 (2004)

\bibitem{Darwiche09}
Darwiche, A.: {Modeling and Reasoning with Bayesian Networks}. Cambridge
  University Press (2009)

\bibitem{Darwiche11}
Darwiche, A.: {SDD: A New Canonical Representation of Propositional Knowledge
  Bases}. In: Proceedings of the Twenty-Second International Joint Conference
  on Artificial Intelligence. pp. 819--826 (2011)

\bibitem{DarwicheM02}
Darwiche, A., Marquis, P.: {A Knowledge Compilation Map}. Journal of Artificial
  Intelligence Research  17(1),  229--264 (2002)

\bibitem{Haken85}
Haken, A.: {The Intractability of Resolution}. Theoretical Computer Science
  39,  297--308 (1985)

\bibitem{Huang10}
Huang, J.: Extended clause learning. Artificial Intelligence  174(15),
  1277--1284 (2010)

\bibitem{HuangD05}
Huang, J., Darwiche, A.: {On Compiling System Models for Faster and More
  Scalable Diagnosis}. In: Proceedings of the Twentieth AAAI Conference on
  Artificial Intelligence. pp. 300--306 (2005)

\bibitem{LagniezM14}
Lagniez, J.M., Marquis, P.: {Preprocessing for Propositional Model Counting}.
  In: Proceedings of the Twenty-Eighth AAAI Conference on Artificial
  Intelligence. pp. 2688--2694 (2014)

\bibitem{LagniezM17}
Lagniez, J.M., Marquis, P.: {On Preprocessing Techniques and Their Impact on
  Propositional Model Counting}. Journal of Automated Reasoning  58(4),
  413--481 (2017)

\bibitem{Manthey14}
Manthey, N.: {Extended Resolution in Modern {SAT} Solving}. In: Joint Automated
  Reasoning Workshop and Deduktionstreffen (2014)

\bibitem{MantheyHB12}
Manthey, N., Heule, M., Biere, A.: {Automated Reencoding of Boolean Formulas}.
  In: Hardware and Software: Verification and Testing - 8th International Haifa
  Verification Conference. pp. 102--117 (2012)

\bibitem{MuiseMB12}
Muise, C., Mcilraith, S.A., Beck, J.C., Hsu, E.: {DSHARP: Fast d-DNNF
  Compilation with sharpSAT}. In: Proceedings of the Twenty-Fifth Canadian
  Conference on Artificial Intelligence. pp. 356--361 (2012)

\bibitem{NamASR04}
Nam, G., Aloul, F.A., Sakallah, K.A., Rutenbar, R.A.: {A Comparative Study of
  Two Boolean Formulations of {FPGA} Detailed Routing Constraints}. IEEE
  Transactions on Computers  53(6),  688--696 (2004)

\bibitem{OztokD15}
Oztok, U., Darwiche, A.: {A Top-Down Compiler for Sentential Decision
  Diagrams}. In: Proceedings of the Twenty-Fourth International Joint
  Conference on Artificial Intelligence. pp. 3141--3148 (2015)

\bibitem{PipatsrisawatD08}
Pipatsrisawat, K., Darwiche, A.: New compilation languages based on structured
  decomposability. In: Proceedings of the Twenty-Third {AAAI} Conference on
  Artificial Intelligence. pp. 517--522 (2008)

\bibitem{RobertsonS84}
Robertson, N., Seymour, P.D.: {Graph Minors. III. Planar Tree-width}. Journal
  of Combinatorial Theory, Series B  36(1),  49--64 (1984)

\bibitem{Robinson65}
Robinson, J.A.: {A Machine-Oriented Logic Based on the Resolution Principle}.
  Journal of the ACM  12(1),  23--41 (1965)

\bibitem{Roth96}
Roth, D.: On the hardness of approximate reasoning. Artificial Intelligence
  82(1-2),  273--302 (1996)

\bibitem{Sauerhoff03}
Sauerhoff, M.: Approximation of boolean functions by combinatorial rectangles.
  Theoretical Computer Science  1--3(301),  45--78 (2003)

\bibitem{SchumannHS10}
Schumann, A., Huang, J., Sachenbacher, M.: {Computing Cost-Optimal Definitely
  Discriminating Tests}. In: Proceedings of the Twenty-Fourth AAAI Conference
  on Artificial Intelligence (2010)

\bibitem{SchumannS10}
Schumann, A., Sachenbacher, M.: {Computing Energy-Optimal Tests Using DNNF
  Graphs}. In: Proceedings of the Twenty-First International Workshop on the
  Principles of Diagnosis (2010)

\bibitem{Tseitin68}
Tseitin, G.S.: {On the Complexity of Derivations in the Propositional
  Calculus}. Studies in Mathematics and Mathematical Logic  Part II,  115--125
  (1968)

\bibitem{Tseitin70}
Tseitin, G.S.: {On the Complexity of Proofs in Propositional Logics}. Seminars
  in Mathematics  8,  466--483 (1970)

\end{thebibliography}

\end{document}